\pdfoutput=1

\documentclass[11pt]{article}

\usepackage[]{ACL2023}

\usepackage{times}
\usepackage{latexsym}
\usepackage[T1]{fontenc}

\usepackage[utf8]{inputenc}

\usepackage{microtype}

\usepackage{inconsolata}
\usepackage{fdsymbol}
\usepackage{multirow}
\usepackage{graphicx}
\usepackage{booktabs}
\usepackage{mathtools}
\usepackage{xspace}
\usepackage{arydshln}
\usepackage{algorithm}
\usepackage{algpseudocode}

\usepackage[nameinlink]{cleveref}
\usepackage{multirow}
\usepackage{enumitem}

\newcommand{\inlinedComment}[2]
{\textcolor{#1}{\small\textbf{#2}}}

\newcommand{\methodnamews}{\text{Self-Improved}}
\newcommand{\methodname}{\methodnamews~}

\crefformat{section}{\S#2#1#3} 
\crefname{algorithm}{Alg.}{Algs.}
\crefformat{subsection}{\S#2#1#3}
\Crefname{equation}{Eq.}{Eqs.}
\Crefname{figure}{Fig.}{Figs.}

\title{Better Language Models of Code through Self-Improvement}

\author{
	Hung Quoc To$^{1}$\thanks{\ \ Equal contribution. Listing order is based on the alphabetical ordering of author surnames.}, \  Nghi D. Q. Bui$^{2}$\footnotemark[1], \ Jin Guo$^{3,4}$,  \ \textbf{Tien N. Nguyen$^{5}$}\\
	$^1$ FPT Software AI Center,
	$^2$Department of Computer Science, Fulbright University, Viet Nam\\
	$^3$School of Computer Science, McGill University, Canada\\
	$^4$Mila - Quebec AI Institute\\
	$^5$School of Engineering and Computer Science , The University of Texas at Dallas, USA\\
	hungtq29@fsoft.com.vn, nghi.bui@fulbright.edu.vn, jguo@cs.mcgill.ca, tien.n.nguyen@utdallas.edu
}

\DeclareMathOperator*{\argmax}{argmax}

\begin{document}
\maketitle

\begin{abstract}


Pre-trained language models for code (PLMCs) have gained attention in recent research. These models are pre-trained on large-scale datasets using multi-modal objectives. However, fine-tuning them requires extensive supervision and is limited by the size of the dataset provided. We aim to improve this issue by proposing a data augmentation framework using knowledge distillation. Our framework utilizes knowledge gained during the pre-training and fine-tuning stage to generate pseudo data, which is then used as training data for the next step. We incorporate this framework into the state-of-the-art  language models, such as CodeT5, CodeBERT, and UnixCoder. The results show that our framework significantly improves PLMCs' performance in sequence-generation tasks, such as code summarization and code generation in the CodeXGLUE benchmark.


\end{abstract}

\section{Introduction}
Pre-trained models for code (PLMCs), such as CodeBERT~\cite{feng-etal-2020-codebert}, PLBART~\cite{ahmad-etal-2021-unified}, CodeT5~\cite{wang-etal-2021-codet5}, UniXCoder~\cite{guo-etal-2022-unixcoder}, and DISCO~\cite{ding-etal-2022-towards}, have become the foundation to solve many practical software engineering tasks such as code summarization, code translation, program repair. 
Those PLMCs, like large language models (LLMs), are typically first pretrained on very large-scale datasets with a variety of multi-modal objectives under a self-supervised training style. They can then be fine-tuned using task-specific datasets in a supervised training style. 


We hypothesise that, while fine-tuned models may not achieve peak performance, PLMCs can produce reasonable outputs that can be regarded as high quality data because they have been pretrained on large scale datasets, and that such data can be leveraged as additional high-quality training data.
Our framework utilizes the self-improvement capability of PLMCs through an simple data augmentation step. This approach is particularly useful for tasks involving code-related sequence generation, such as code summarization and code generation. Our method involves fine-tuning a PLMC on a downstream dataset, allowing the model to gain knowledge about the task. The model then generates pseudo outputs, which are used in conjunction with the original training data to train for the next epoch. Our framework is similar to sequence-level knowledge distillation~\cite{kim-rush-2016-sequence}, but our approach focuses on improving model performance without compressing the model by utilizing the same technique.


Our empirical evaluation results show that our framework significantly improves the  state-of-the-arts PLMCs, including CodeBERT, CodeT5, UniXCoder with significant margins. In short, we summarize our contributions as follows.

 \begin{itemize}[leftmargin=*]
 \itemsep0em
     \item We present a simple self-improvement framework and show how it can be easily adapted to PLMCs for the task of code-related sequence generation.
    \item We conduct extensive evaluation on two  tasks: code summarization and code generation, and compare it with the well-known, state-of-the-art PLMCs. The results show that our framework consistently improvesover all PLMCs by a significant margin in those tasks.
     \item We provide  analysis and  explanations on how utilizing a simple framework consistently improves the performance of PLMCs.
 \end{itemize}

\begin{figure}
 \centerline{\includegraphics[width=1\linewidth]{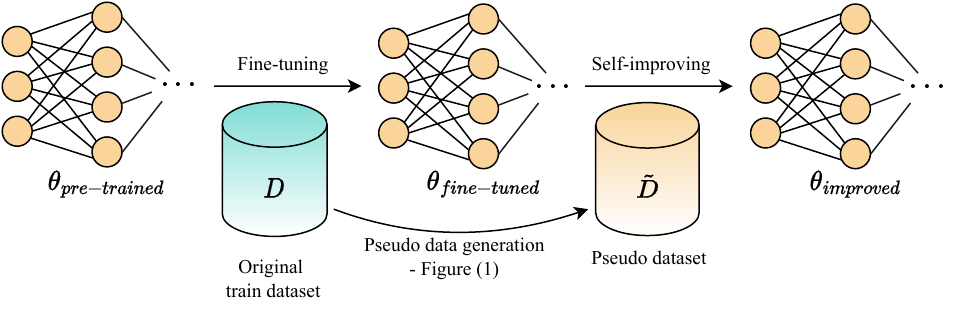}}
\caption{Overall training pipeline.}\label{figure:method1}
\vspace{-1em}
\end{figure}

\begin{figure}
 \centerline{\hspace{0.7em} \includegraphics[width=1.0\linewidth]{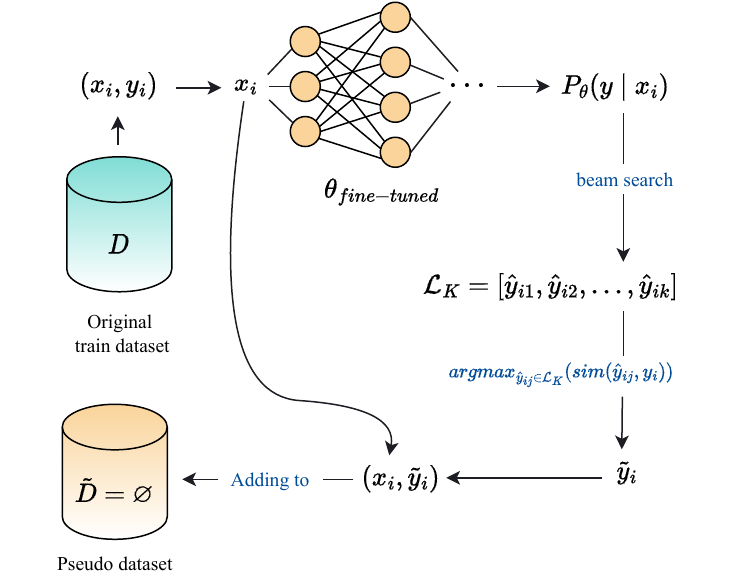}}
\caption{Demonstrating the process of generating pseudo dataset in our work.}\label{figure:method2}
\vspace{-1em}
\end{figure}

\section{Related Work}

\paragraph{Exposure bias and hallucination in Sequence Generation Tasks}
The exposure bias problem is regarded as the difference between the training and inference phases for auto-regressive sequence generation models. Previous work has attempted to reduce exposure bias in training phase~\cite{bengio2015scheduled,ranzato2015sequence,wiseman2016sequence, wang2020exposure}.
In the sense that our self-improvement step involves training model on its own prediction, the exposure bias is close to our approach.
\paragraph{Code understanding and generation}
Code learning problems have recently emerged as one of the primary tasks for assessing the capability of language models. Most recent code models are pretrained on multi-modal objectives before being fine-tuned on specific downstream tasks ~\cite{feng-etal-2020-codebert, ahmad-etal-2021-unified,wang-etal-2021-codet5,guo-etal-2022-unixcoder,ding-etal-2022-towards}.
\paragraph{Knowledge Distillation}
Knowledge distillation is the process of transferring knowledge from a large unwieldy model or set of models to a single smaller model that can be practically deployed under real-world constraints, and such smaller model can usually keep the same performance or even better than the original model
~\cite{44873,kim-rush-2016-sequence,DBLP:journals/corr/abs-2002-10957,DBLP:journals/corr/abs-2006-10029,DBLP:journals/corr/abs-2106-04563}. We perform an additional self-improvement step to improve the original model without using  external resources, our work is relevant to knowledge distillation.

\section{Method}

\begin{table*} [t]
\begin{center}
\resizebox{0.99\textwidth}{!}{
\begin{tabular}{p{12cm}p{7cm}}
\begin{minipage}{0.75\textwidth}
\resizebox{\textwidth}{!}{
\begin{tabular}{lc c lc c c c c cc}
\toprule
Models & Beam sizes & Methods & Ruby & JavaScript & Go & Python & Java & PHP & Overall \\
\midrule
    RoBERTa & 10 &\cite{DBLP:journals/corr/abs-1907-11692} & 11.17 & 11.90 & 17.72 & 18.14 & 16.47 & 24.02 & 16.57 \\
PLBART & 10 &\cite{ahmad-etal-2021-unified}& 14.11 & 15.56 & 18.91 & 19.30 & 18.45 & 23.58 & 18.32 \\
PolyglotCodeBERT & 10 &\cite{DBLP:journals/corr/abs-2112-02043} & 14.75 & 15.80 & 18.77 & 18.71 & 20.11 & 26.23 & 19.06 \\
\midrule
CodeBERT & 1 & Baseline          & 12.04 & 14.73 & 17.58 & 18.47 & 17.62 & 23.44 & 17.31   \\
         &   & \methodname               & \textbf{12.40} & \textbf{15.44} & \textbf{18.52} & \textbf{19.09} & \textbf{18.31} & \textbf{24.46} & \textbf{18.04} \\
\cdashline{2-10}
         & 5 & Baseline          & 12.31 & 15.76 & 18.12 & 19.09 & 18.37 & 24.85 & 18.08 \\
         &   & \methodname               & \textbf{12.55} & \textbf{16.41} & \textbf{18.69} & \textbf{19.50} & \textbf{18.88} & \textbf{25.21} & \textbf{18.54} \\
\cdashline{2-10}
         & 10 & Baseline          & 12.22 & 15.78 & 18.01 & 19.09 & 18.42 & 25.06 & 18.10\\
         &    & \methodname               & \textbf{12.52} & \textbf{16.39} & \textbf{18.66} & \textbf{19.50} & \textbf{18.92} & \textbf{25.28} & \textbf{18.54} \\
\midrule
CodeT5 (base) & 1 & Baseline          & 14.80 & 15.57 & 19.57 & 19.86 & 19.87 & 25.33 & 19.17 \\
       &   & \methodname & \textbf{15.46} & \textbf{16.22} & \textbf{20.12} & \textbf{20.36} & \textbf{20.25} & \textbf{26.25} & \textbf{19.78} \\
\cdashline{2-10}
       & 5 & Baseline          & 15.23 & 16.18 & 19.95 & 20.42 & 20.26 & 26.11 & 19.69 \\
       &   & \methodname & \textbf{15.60} & \textbf{16.51} & \textbf{20.26} & \textbf{20.59} & \textbf{20.44} & \textbf{26.46} & \textbf{19.97} \\
\cdashline{2-10}
       & 10 & Baseline          & 15.29 & 16.18 & 19.95 & 20.42 & 20.26 & 26.10 & 19.70 \\
       &    & \methodname & \textbf{15.70} & \textbf{16.47} & \textbf{20.28} & \textbf{20.58} & \textbf{20.45} & \textbf{26.58} & \textbf{20.00} \\
\midrule
    UniXCoder (base) & 1 & Baseline          & 14.83 & 15.39 & 18.95 & 18.66 & 19.37 & 24.99 & 18.70 \\
       &   & \methodname & \textbf{15.36} & \textbf{15.83} & \textbf{19.42} & \textbf{19.13} & \textbf{20.04} & \textbf{26.05} & \textbf{19.31} \\ 
\cdashline{2-10}
       & 5 & Baseline          & 15.17 & 15.93 & 19.52 & 19.25 & 20.18 & 26.45 & 19.42 \\
       &   & \methodname & \textbf{15.37} & \textbf{15.95} & \textbf{19.73} & \textbf{19.55} & \textbf{20.45} & \textbf{26.69} & \textbf{19.62} \\
\cdashline{2-10}
       & 10 & Baseline          & 14.74 & 15.84 & 19.31 & 19.12 & 20.11 & 26.75 & 19.31 \\
       &    & \methodname & \textbf{15.37} & \textbf{15.96} & \textbf{19.73} & \textbf{19.56} & \textbf{20.44} & \textbf{26.79 }& \textbf{19.63} \\

\bottomrule
\end{tabular}
}

\end{minipage}
\end{tabular}
}
\captionsetup{type=table}
\vspace{-0.5em}  
\caption{
    Results on code summarization evaluated with smoothed BLUE-4.
The ``Overall'' column presents average scores over six programming languages.
The bolded numbers represent the best scores (higher is better) when comparing between Baseline and \methodname for each model and each value of beam size.
}\label{table:summarize}
\end{center}
\vspace{-1em}
\end{table*}


Our method utilizes three different sets of model parameters: $\theta_{pre-trained}$, $\theta_{fine-tuned}$, and $\theta_{improved}$. Each corresponds to the stage of the model parameters after pre-trained, fine-tuned, and self-improved, respectively. The model generates tokens auto-regressively, token-by-token. 

Usually, models are pretrained on large scale corpora, resulting in a pre-trained checkpoint $\theta_{pre-trained}$. These pre-trained models are then fine-tuned on a specific downstream dataset $D$ using a supervised-learning approach, resulting in a set of fine-tuned parameters $\theta_{fine-tuned}$. Our investigation revealed that model performance can be further improved if we continue to fine-tuned these parameters on an augmented version of $D$. 
As depicted in Figure~\ref{figure:method1}, our proposal for self-improvement is the final step in the overall training flow.  Specifically, we propose a data augmentation process and an extra fine-tuning step in addition to the pre-training and fine-tuning paradigm. The process of augmenting the dataset is illustrated in Figure~\ref{figure:method2}. We also give a detailed algorithm for this process in the Appendix.
For each training pair of sequences $(x_i, y_i)$ in the train dataset $D$, we first use beam search to generate a list of K-best predictions $L_{K}$. This list contains $k$ predictions, where $k$ is the beam size.

We then evaluate the similarity of each prediction $\hat{y}_{ij}$ and its corresponding ground truth sequence $y_i$ using a similarity function $sim$ based on BLEU score. The best prediction with highest similarity is then selected $\tilde{y}_i = \argmax_{\hat{y}_{ij} \in \mathcal{L}_K}(sim(\hat{y}_{ij}, y_i))$. In the last step, we add the pair of sequences $(x_i, \tilde{y}_i)$ into a new empty dataset $\tilde{D}$. We call this new dataset the {\em augmented dataset or pseudo dataset} interchangeably in the rest of the paper.
The next step requires fine-tuning $\theta_{fine-tuned}$ on $\tilde{D}$ until convergence. We call this new stage of model parameters $\theta_{improved}$.
Note that the index $j$ in $\hat{y}_{ij}$ denotes the $j^{th}$ prediction in the beam, not the $j^{th}$ token of the predicted sequence. Additionally, only train dataset $D$ is augmented, while the validation and test dataset are kept unchanged for evaluation purpose.

\section{Experimental Setup}

Our goal is to show that for any of the fine-tuned model for a sequence generation task (F-PLMC), after applying our self-improvement method (S-PLMC), the result improves.


\paragraph{Dataset and Downstream Tasks}
To achieve our goal of enhancing the code-related sequence generation task, we selected code summarization and code generation as our experimental areas. To evaluate these tasks, we utilized the CodeXGLUE benchmark~\cite{DBLP:journals/corr/abs-2102-04664}, which comprises various datasets for various code understanding and code generation tasks. Specifically, we utilized the code summarization and code generation datasets from CodeXGLUE and disregarded the other ones. The statistics for each dataset is reported in Appendix.

\paragraph{Baseline Models}
We chose CodeBERT~\cite{feng-etal-2020-codebert}, CodeT5~\cite{wang-etal-2021-codet5}, and UniXCoder~\cite{guo-etal-2022-unixcoder} as baseline models. Each model represents a distinct neural architecture style. 
CodeBERT abides to the Bidirectional Transformer architecture, which is a well-known PLMCs, despite the fact that it may not produce the best results for the tasks in CodeXGLUE. CodeT5 and UniXCoder are the two PLMCs that achieve state-of-the-arts performance on the CodeXGLUE benchmark. CodeT5 is an encoder-decoder architecture that follows the Seq2Seq learning style by following T5.
UniXCoder, on the other hand, is a unified language model. It can behave as an encoder-only, decoder-only, or encoder-decoder model by modifying the masked attention matrices inside each transformer layer. Note that while CodeT5 and UniXCoder are proposed to address both code summarization and code generation, CodeBERT only considers the first problem in there paper. So we only consider CodeBERT for code summarization in our work.

\paragraph{Evaluation Metric}
We follow CodeXGLUE benchmark in employing evaluation metrics for each task.
Smoothed BLEU-4~\cite{lin-och-2004-orange} is used as the evaluation metric for code summarization.
For code generation, smoothed BLEU-4, CodeBLEU~\cite{DBLP:journals/corr/abs-2009-10297}, and exact match (EM) are employed.
We aim to improve all of these metrics after apply our self-improvement method into the PLMCs.

\paragraph{Implementation}
We carefully selected the checkpoints for CodeBERT~\footnote{\url{https://github.com/microsoft/CodeBERT/tree/master/CodeBERT}}, CodeT5~\footnote{\url{https://github.com/salesforce/CodeT5}}, and UniXCoder~\footnote{\url{https://github.com/microsoft/CodeBERT/tree/master/UniXcoder}} based on the corresponding tasks. It is important to note that not all of these methods have released fine-tuned checkpoints. CodeT5 stands out as the only model to have released checkpoints for code summarization and code generation tasks. Conversely, CodeBERT and UniXCoder only offer training scripts. When checkpoints were not available, we employed the provided training scripts to fine-tune the pretrained models until we obtained results comparable to those reported in the original research.
Additionally, CodeT5 and UniXCoder may have different checkpoint options with varying model sizes, such as \textit{small}, \textit{base}, and \textit{large}. We selected the \textit{base} version for both CodeT5 and UniXCoder.

\section{Evaluation Results}

\begin{table} [t]
\centering
\resizebox{\columnwidth}{!}{
\begin{tabular}{lc c c c c c}
\toprule
    Models & Beam sizes & Methods & EM & BLEU & CodeBLEU \\ 
\midrule
    CodeGPT & 10 & \cite{DBLP:journals/corr/abs-2102-04664} & 20.10 & 32.79 & 35.98\\
    PLBART & 10 & \cite{ahmad-etal-2021-unified} & 18.75 & 36.69 & 38.52 \\
\midrule
    CodeT5 (base) & 1 & Baseline & 21.75 & 39.00 & 41.64 \\
                  &  & \methodname & \textbf{22.40} & \textbf{39.75} & \textbf{42.14} \\
\cdashline{2-7}
       & 5 & Baseline & 21.10 & 40.67 & 43.59\\
       && \methodname & \textbf{22.40} & \textbf{41.61} & \textbf{44.06} \\
\cdashline{2-7}
       & 10 & Baseline & 22.10 & 39.59 & 43.78 \\
       && \methodname & \textbf{22.35} & \textbf{41.85} & \textbf{44.49} \\
\midrule
    UniXCoder (base) & 1 &  Baseline    & 21.50 & 38.28 & 40.85 \\
              &   & \methodname & \textbf{22.10} & \textbf{38.56} & \textbf{40.96} \\
\cdashline{2-7}
       & 3 & Baseline & 22.05 & 37.53 & 40.11 \\
       &   & \methodname & \textbf{22.30} & \textbf{37.88} & \textbf{40.42} \\
\cdashline{2-7}
       & 5 & Baseline & 22.00 & 37.18 & 39.78 \\
       &   & \methodname & \textbf{22.30} & \textbf{37.49} & \textbf{40.05} \\
\bottomrule
\end{tabular}
}

\vspace{-0.5em}  
\caption{
    Results on code generation evaluated with EM, BLEU, and CodeBLEU.
The bolded numbers represent the best scores (higher is better for all metrics) when comparing between Baseline and \methodname for each model and each value of beam size.
}\label{table:concode}
\vspace{-1em}
\end{table}

The results of our code summarization task are presented in Table~\ref{table:summarize}. The "Beam sizes" column indicates the beam size used in the beam search algorithm, while the "Methods" column indicates whether or not our self-improved algorithm was utilized. We also included other models as references to compare the relative improvement of our model.
On average, we observed an average of 0.76 BLUE score increase in performance across all languages. This improvement was consistent across various beam sizes (1, 5, 10), which confirms the effectiveness of our self-improved approach across a wide range of PLMCs.
When comparing our model to other strong baselines, we found that our method improved the performance of CodeBERT for JavaScript from 15.78 to 16.39, surpassing the performance of PolyglotCodeBERT (15.80). This highlights the benefit of our self-improved method in improving weak models.
The results of our code generation study are presented in Table~\ref{table:concode},   the performance increase by 0.81 BLUE scores on average. When using EM and CodeBLEU, the improvement also increases consistently.


\section{Ablation Study}
\label{sec:discussion}
\begin{figure}
 \centerline{\includegraphics[width=1\linewidth]{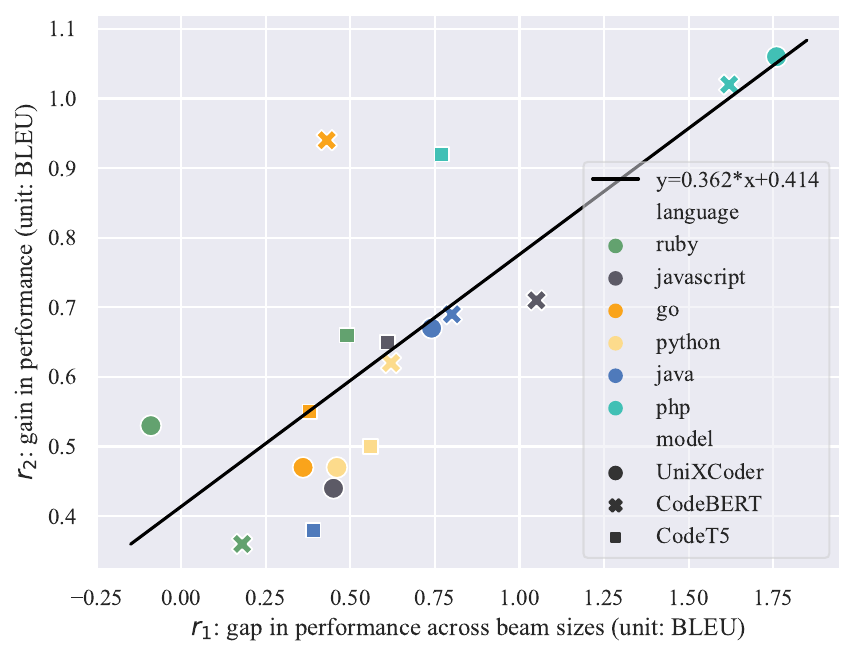}}
\caption{
Scatter plot visualizing 
performance gap (in BLEU score) infered by different beam sizes (i.e 10 and 1) of $\theta_{fine-tuned}$
vs. 
performance gained (in BLEU score) by $\theta_{improved}$ infered with beam size of 1
}\label{figure:plot}
\vspace{-1em}
\end{figure}

In this section, we examine the factors that influence the improvement achieved by $\theta_{improved}$ as compared to $\theta_{fine-tuned}$ through code summarization. We define $r_1$ as the difference in performance measured by BLEU between inferencing with a beam size of 10 and inferencing with a beam size of 1. Additionally, we define $r_2$ as the improvement in BLEU when inferencing with the same beam size of 1 between $\theta_{fine-tuned}$ and $\theta_{improved}$. By evaluating these values across a variety of beam sizes and programming languages in the code summarization dataset, we are able to visualize the results in Figure~\ref{figure:plot}. Additionally, we have calculated the Pearson Correlation score, which is 0.77, indicating a strong correlation between $r_1$ and $r_2$. Our analysis demonstrates that a larger $r_1$ is correlated with a better $r_2$, suggesting that our method is more likely to yield better overall performance when $r_1$ is large. We believe this insight is a crucial finding as it provides a simple indicator of the model's fully trained capability.


\section{Conclusion}
We introduced a self-improvement technique as a final fine-tuning step to enhance model performance. Our experiments showed that this method, when applied to popular pre-trained code models (CodeBERT, CodeT5, and UniXCoder), significantly improves performance on code summarization and code generation tasks. We also provided insights on when this method is most effective in improving PLMCs. We intend to implement our technique in larger-scale models and other tasks, and believe it is an efficient way to optimize the capabilities of any code language model without the need for extensive architecture modifications or large-scale dataset assembly. We leave all of these investigations for the future.

\newpage
\section*{Limitations}
We discuss a few limitations of our work.
One limitation of \methodname is its complexity in usage. The process of generating pseudo data involves generating predictions for the entire training dataset with a large beam size, resulting in a time complexity of $O(n k)$, where $n$ is the train dataset size and $k$ is the beam size. Additionally, the fine-tuning step to derive $\theta_{improved}$ also adds a significant amount of computational complexity. This limitation is discussed in Section~\ref{sec:discussion} to weigh the performance benefits of our method against the computational sacrifices. Another limitation is that \methodname has only been applied to encoder-decoder models in this work. However, it is also applicable to other types of auto-regressive models, including encoder-only models, which are commonly used for tasks such as code completion~\cite{radford2019language,DBLP:journals/corr/abs-2102-04664,guo-etal-2022-unixcoder}. A few models can be named are GPT models~\cite{radford2019language,DBLP:journals/corr/abs-2005-14165}, CodeX~\cite{DBLP:journals/corr/abs-2107-03374}, CodeGen~\cite{https://doi.org/10.48550/arxiv.2203.13474}, etc. Further research into these applications is left for future work.

\bibliography{acl2023}
\bibliographystyle{acl_natbib}

\appendix
\section{Algorithms}
\renewcommand{\algorithmicrequire}{\textbf{Input:}}
\renewcommand{\algorithmicensure}{\textbf{Output:}}
\algnewcommand\RETURN{\State \algorithmicreturn}%

\algnewcommand\algorithmicforeach{\textbf{for each}}
\algdef{S}[FOR]{ForEach}[1]{\algorithmicforeach\ #1\ \algorithmicdo}

\begin{algorithm}[H]
\caption{Pseudo Data Generation}
\begin{algorithmic}[1]
    \Require{
        \Statex \textbullet~ $\theta_{fine-tuned}$, the fine-tuned model checkpoint on a specific task $T\in\{$code summarization, code generation, etc. $\}$.
        \Statex \textbullet~ $D=\{(x_i, y_i)\mid i=\overline{1,n}\}$, the train dataset on which the $\theta_{fine-tuned}$ is fine-tuned.
        \Statex \textbullet~ $B_{k}$ denotes the $beam search$ algorithm with beam size of $k$. It returns a list of $k$ best sequences as prediction.
}
    \Ensure{
        \Statex \textbullet~ Pseudo dataset $\tilde{D}$
    }
\Procedure{GeneratingPseudoData}{}
\State $\tilde{D}\leftarrow \varnothing$
\ForEach {datapoint $(x_i, y_i) \in D$}:
    \State $\mathcal{L}_K \leftarrow B_{k}(P_{\theta_{fine-tuned}}(y \mid  x_i))$
    \State In other words, $\mathcal{L}_{K} = [\hat{y}_{i1}, \hat{y}_{i2}, ..., \hat{y}_{ik}]$
    \State $\tilde{y}_i \leftarrow \argmax\limits_{\hat{y}_{ij} \in \mathcal{L}_{K}}(sim(\hat{y}_{ij}, y_i))$
	\State Adding $(x_i, \tilde{y}_i) \rightarrow \tilde{D}$
\EndFor
    \RETURN{} $\tilde{D}$ 
\EndProcedure
\end{algorithmic}
\end{algorithm}


\section{Hyperparameter selection}

We keep all the values of hyperparameters as in the training script for each model, except that we increase the batch size in order to utilized completely memory of a NVIDIA A100 80GB.
\section{Data statistics}
\begin{table} [t]
\captionsetup{}
\centering
\resizebox{\columnwidth}{!}{
\begin{tabular}{lc c c c }
\toprule
Programming Language & Training&Dev& Test \\
\midrule
Python & 251,820 & 13,914 & 14,918 \\
\midrule
PHP & 241,241 & 12,982,& 14,014 \\
\midrule
Go & 167,288 & 7,325 & 8,122 \\
\midrule
Java & 164,923 & 5,183 & 10,955 \\
\midrule
JavaScript & 58,025& 3,885 & 3,291 \\
\midrule
Ruby & 24,927& 1,400 & 1,261 \\
\bottomrule
\end{tabular}
}

\vspace{-0.5em}  
\caption{
Statictisc of data used in code summarization
    }\label{table:stat_csn}
\end{table}

\begin{table} [t]
\captionsetup{}
\centering
\resizebox{0.2\textwidth}{!}{
\begin{tabular}{lc c}
\toprule
Split & \#Examples \\
\midrule
Train & 100,000 \\
\midrule
Dev & 2,000 \\
\midrule
Test & 2,000 \\
\bottomrule
\end{tabular}
}

\vspace{-0.5em}  
\caption{
Statictisc of data used in code generation
    }\label{table:stat_concode}
\vspace{-1em}
\end{table}

We include the statstics for the data used in our experiments. We access directly to CodeXGLUE page\footnote{\url{https://github.com/microsoft/CodeXGLUE}} for downloading data. CodeXGLUE gathers datasets from multiple sources for each downstream task.
For code summarization, the CodeSearchNet~\cite{DBLP:journals/corr/abs-1909-09436} are used with the number of examples for each train/dev/test are reported in Table~\ref{table:stat_csn}.
The dataset comprise the code-text pairs in 6 programming languages.
CONCODE~\cite{DBLP:journals/corr/abs-1808-09588} dataset is employed as the benchmark for code generatiom with  statistics reported in Table~\ref{table:stat_concode}.
It contains pairs of Java member function and natural language description Java language.

\end{document}